\documentclass[letterpaper, 10 pt, journal, twoside]{IEEEtran}  
\IEEEoverridecommandlockouts                              

\usepackage{epsfig} 
\usepackage{times} 
\usepackage{hyperref}
\usepackage{mathtools}
\usepackage{xcolor}

\usepackage{amsmath} 
\usepackage{amssymb}  
\usepackage{mathrsfs}
\usepackage{comment} 
\usepackage{algorithm,algorithmic}
\begin{document}

\title{Dynamic and Versatile Humanoid Walking via Embedding 3D Actuated SLIP Model with Hybrid LIP Based Stepping}
    \author{Xiaobin Xiong and Aaron D. Ames
\thanks{Manuscript received: Feb. 24, 2020; Revised May 22, 2020; Accepted June 20, 2020.}
\thanks{This paper was recommended for publication by Editor Abderrahmane Kheddar upon evaluation of the Associate Editor and Reviewers' comments.
This work was supported by Amazon-Caltech Fellowship in AI. \textit{(Corresponding author:
Xiaobin Xiong.)}} 
\thanks{The authors are with the Department of Mechanical and Civil Engineering, California Institute of Technology, Pasadena, CA 91125
        {\tt\footnotesize xxiong@caltech.edu}, {\tt\footnotesize ames@caltech.edu}}%
        \thanks{Digital Object Identifier (DOI): see top of this page.}
 }

\maketitle

\markboth{IEEE Robotics and Automation Letters. Preprint Version. Accepted June, 2020}
{Xiong \MakeLowercase{\textit{et al.}}: Dynamic and Versatile Humanoid Walking} 

\begin{abstract}
In this paper, we propose an efficient approach to generate dynamic and versatile humanoid walking with non-constant center of mass (COM) height. We exploit the benefits of using reduced order models (ROMs) and stepping control to generate dynamic and versatile walking motion. Specifically, we apply the stepping controller based on the Hybrid Linear Inverted Pendulum Model (H-LIP) to perturb a periodic walking motion of a 3D actuated Spring Loaded Inverted Pendulum (3D-aSLIP), which yields versatile walking behaviors of the 3D-aSLIP, including various 3D periodic walking, fixed location tracking, and global trajectory tracking. The 3D-aSLIP walking is then embedded on the fully-actuated humanoid via the task space control on the COM dynamics and ground reaction forces. The proposed approach is realized on the robot model of Atlas in simulation, wherein versatile dynamic motions are generated.  
\end{abstract}


\IEEEpeerreviewmaketitle

\section{INTRODUCTION}

\IEEEPARstart{P}{lanning} dynamic and versatile humanoid walking is a non-trivial task. Traditionally, a lot of research has been focused on generating the desired center of mass (COM) motion via the Linear Inverted Pendulum (LIP) model \cite{kajita2003biped, feng2016robust}, where the zero moment point (ZMP) \cite{vukobratovic2004zero} inside the support polygon describes the feasibility of the motion. Additionally, centroidal momentum-based controls \cite{dai2014whole, wensing2013generation} have also been proposed to better capture the additional rotational dynamics that contributes to the dynamics of the ZMP. Those approaches oftentimes promote fast planning but with relatively conservative COM dynamics, i.e., the walking has to be with a constant COM height due to the LIP assumption. 

As a counterpart, researchers in the underactuated walking community propose dynamic walking generation based on the full-dimensional model of the robot \cite{grizzle2014models, westervelt2003hybrid,hereid20163d}. Large and non-convex optimization problems \cite{hereid20163d} are formulated to optimize the desired motion subject to the physical constraints. The generated walking motion can be dynamic in the sense that there is no restriction on the COM height. However, the optimization problems are computationally expensive to solve, and their feasibilities (existence of solutions) are subject to fine-tunings on the cost and constraint functions.

One approach to remove the COM height constraint is embedding the COM dynamics \cite{wensing2013generation, Garofalo2012WalkingCO} of the walking of the canonical Spring Loaded Inverted Pendulum (SLIP) model \cite{geyer2006compliant,full1999templates} on the humanoid via optimization-based control. The periodic motion of the SLIP has to be numerically identified via simulation, and non-periodic motion has not been well studied due to its nonlinear dynamics. Thus, it is not efficient to generate versatile walking using this approach. 
\begin{figure}[t]
      \centering
      \includegraphics[width = 1\columnwidth]{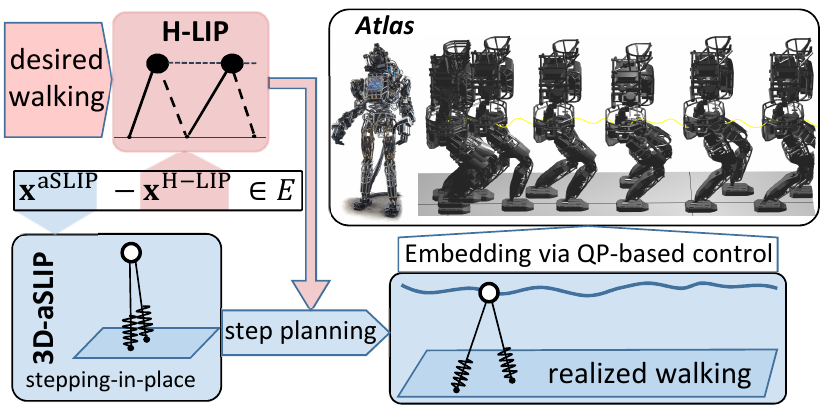}
      \caption{ Overview of the approach.}
      \label{fig:overview}
\end{figure}

To efficiently generate dynamic and versatile walking, we propose walking generation via a 3D actuated Spring Loaded Inverted Pendulum (3D-aSLIP) model with the application of a stepping controller based on the Hybrid-Linear Inverted Pendulum (H-LIP) model (see Fig. \ref{fig:overview}). The 3D-aSLIP is an extension of the planar aSLIP model \cite{xiong2018coupling} to the 3D scenario. The H-LIP with its stabilization \cite{xiong2019IrosStepping} is a formal adaptation of the LIP in \cite{kajita2002realtime}, which is stabilized via changing the step size.

Desired walking behaviors are directly described on the H-LIP, and the stepping controller is designed based on the H-LIP to perturb the step size of a stepping-in-place walking of the aSLIP to behave similarly to the H-LIP walking. The mechanism behind the stepping is the treatment of the step size as the control input to the step-to-step (S2S) dynamics of the walking of the 3D-aSLIP; the S2S of the H-LIP approximates the S2S of the 3D-aSLIP. The stepping controller keeps the difference of the states of the S2S between the 3D-aSLIP and the H-LIP in a minimum disturbance invariant set. Thus, the walking of the 3D-aSLIP behaves similarly to that of the H-LIP. The generated 3D-aSLIP walking is embedded on the humanoid via a quadratic program (QP) based controller. 

The generated walking is first featured to be dynamic (less conservative) with the variation on the COM height. Compared to existing methods for generating the COM height variation in \cite{griffin2018straight,brasseur2015robust} via the kinematic structuring on the leg extension, the aSLIP embedding provides coherent trajectories at the dynamics level. Moreover, the versatile motion generation via the H-LIP based stepping is highly efficient. Trajectory optimization is only required to perform once on the 3D-aSLIP to generate a periodic stepping-in-place motion. The stepping controller is solved in closed-form for periodic walking or via fast solvable quadratic programs for versatile walking tasks. The computation request is minimum at the planning level.

We use the robot model of Atlas to evaluate the proposed approach for realizing versatile walking motions, including periodic walking, fixed location tracking, and global trajectory tracking. Those behaviors are generated on the 3D-aSLIP and then embedded on Atlas successfully. We start by presenting the 3D-aSLIP in section \ref{section:aSLIP} and the H-LIP in section \ref{section:LIP}. The versatile walking generation via stepping is described in section \ref{section:walking}. Section \ref{section:robot} describes the embedding approach and walking results. Lastly, we finish with discussions in section \ref{section:discussion} and conclude with section \ref{section:conclusion}.

\section{3D Actuated SLIP Model}
\label{section:aSLIP}
In this section, we present the model of the 3D actuated Spring Loaded Inverted Pendulum (3D-aSLIP), the step-to-step (S2S) dynamics formulation, and the overview of the stepping controller based on the approximation of the S2S dynamics for generating versatile walking behaviors. 

\subsection{Dynamics Model of Walking}
\begin{figure}[t]
      \centering
      \includegraphics[width = 1\columnwidth]{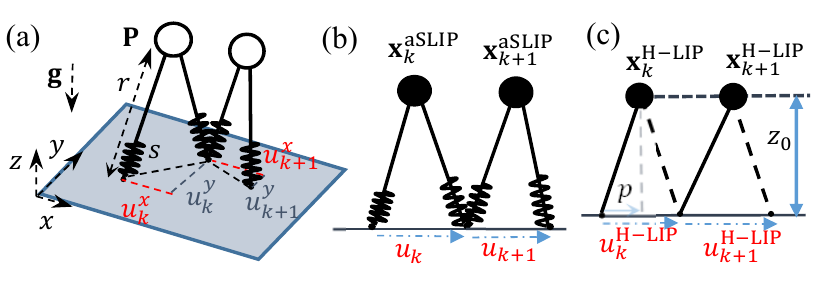}
      \caption{The 3D-aSLIP model (a), planar illustration of the step-to-step (S2S) dynamics of the aSLIP (b) and the S2S dynamics of the H-LIP (c). }
      \label{fig:aSLIP}
\end{figure}

The 3D-aSLIP is consisted by a point mass and two spring-loaded legs (see Fig. \ref{fig:aSLIP} (a)). In the later, we drop the '3D' acronym for conciseness. Depending the number of contacts with the ground, the walking is divided into two alternating phases, i.e., Double Support Phase (DSP) and Single Support Phase (SSP). The dynamics of the point mass is: 
\begin{align}
m \ddot{\mathbf{P}} = \textstyle\sum \mathbf{F} + m \mathbf{g},
\end{align}
where $m$ is the mass of the aSLIP, $\mathbf{P} = [p_x, p_y, p_z]^T$ is the position of the point mass, $\mathbf{F} = \mathbf{F}_{\text{L/R}}$ are the leg forces on the \textit{left} and \textit{right} legs, and $\mathbf{g}$ is the gravitational vector. The leg force is zero when the leg is not in contact with the ground. The magnitude of the leg force is: 
$   | \mathbf{F}_{\text{L/R}}| = \text{K}_s s + \text{D}_s \dot{s},
$ where $s$ is the spring deformation, and $\text{K}_s$ and $ \text{D}_s$ are the stiffness and damping of the leg spring, respectively. 

\textbf{Actuation:} We add actuation on the leg to enable leg extension and retraction. This is inspired by the physical design of the bipedal robot Cassie \cite{xiong2018coupling}. Its leg has leaf springs under-actuated by the actuation of the motor joints which are above the springs. One can view the leg as a serial-elastic actuator, where the springs compose the elastic element at the output of the leg, and the actuation is changing the uncompressed leg length  $L$. Mathematically, as a result of simplification, we define the actuation as $\tau^{\ddot{L}} = \ddot{L}$. The actual leg length is $r = L - s$. Additionally, the step size $u = [u_x, u_y]^T$ is another input to the system. Since the legs have no inertia, the step size can be set directly. In the literature, this has also been equivalently referred to as the touch-down angle control \cite{raibert1986legged, vejdani2015touch} of the swing leg. 

\textbf{Comparison to the canonical SLIP:} The canonical SLIP \cite{geyer2006compliant} typically has no actuation or damping in the leg, thus the system energy is conserved. The purpose of adding actuation and damping is to create mechanisms to inject and dissipate energy so that the system energy can be changed freely, which facilities versatile walking generation. 

\textbf{Impact and Liftoff Model:} When the swing leg strikes the ground, the swing foot velocity has a discrete jump. We assume the discrete jump is on the spring deformation rate $\dot{s}$, whereas the velocity of the uncompressed leg $\dot{L}$ remains unchanged. Due to the damping, the discontinuity of $\dot{s}$ creates a discontinuity on the leg force on the swing leg. The stance leg lifts off from the ground when its leg force crosses zero, and then the spring deformation is set to 0.

\subsection{Trajectory Optimization for Periodic Walking}
The periodic walking trajectory of the 3D-aSLIP can be generated via trajectory optimization \cite{hereid20163d}. Generating versatile dynamic behaviors of walking requires formulating and solving individual optimizations. This process can be tedious. Additionally, it typically requires careful tunings on the cost and additional constraint functions to avoid infeasibility.

In the previous work \cite{xiong2019IrosStepping} and this work, we only rely on trajectory optimization to generate a single periodic walking, e.g., a stepping-in-place walking, and then utilize stepping control to generate a wide family of walking, which eliminates the need of formulating and solving extensive trajectory optimization problems on the aSLIP, or any trajectory optimization on the robot model that embeds the aSLIP dynamics. 

The trajectory optimization for generating the periodic stepping-in-place is formulated using direct collocation \cite{hereid20163d} as,
\begin{align}
\label{SLIPopt}
z^* = \underset{z}
 {\text{argmin}}   & \sum_{i = 1}^N \frac{\Delta t}{2} ({{\ddot{L}}_\text{L}}^{i^2} + {{\ddot{L}}_\text{R}}^{i^2}),\\
\text{s.t.}
\quad &  \mathbf{c}_{\text{min}} \leq \mathbf{c}(z) \leq \mathbf{c}_{\text{max}}, \nonumber
\end{align}
where $\Delta t$ is the time discretization, $N$ is the number of nodes, $z$ represents all the optimization variables, and $\mathbf{c}(z)$ includes all the constraints: friction cone, non-negative normal force, behavior specifications, etc.
The leg length is controlled by: 
\begin{equation}
\label{leglengthActuation}
    \tau^{\ddot{L}} = \ddot{L}^{\text{des}}(t) + K_p (L - L^{\text{des}}(t)) + K_d (\dot{L} - \dot{L}^{\text{des}}(t)),
\end{equation}
 where $L^{\text{des}}$ is the optimized periodic trajectory of the leg length, and $K_p, K_d$ are the feedback gains. Periodically actuating the leg length to the desired trajectory, the aSLIP performs the optimized periodic stepping-in-place. If the step size is perturbed, the aSLIP may fall over or evolve to different behaviors \cite{xiong2019IrosStepping}. In the following, we explain at the high-level the approach of stepping control on the step-to-step (S2S) dynamics with the integration of the periodic leg length actuation, which perturbs the stepping-in-place to versatile walking behaviors. Note that the same leg length trajectory drives the walking to have same step frequency, thus the feasible walking velocity is finite and bounded by its feasible step size\footnote{The average walking velocity is $\frac{u}{T}$ ($T$ is the period of walking) and the maximum step size is $r_\text{L} +r_\text{R}$.}. 
 
\subsection{S2S Dynamics:} 
Assuming an existing periodic walking, the hybrid dynamics of the aSLIP can be described by a discrete dynamical system. Considering the instant when the swing toe strikes the ground, the system state transverses the switching surface, defined by:
\begin{equation}
\label{swithingsurface}
\mathbf{S} = \{ (q, \dot{q}) \in \mathbb{R}^n | p^z_{\text{swing}}(q, \dot{q}) = 0, v^z_{\text{swing}}(q, \dot{q}) < 0 \}, \nonumber
\end{equation}
where $(q, \dot{q})$ describes all the states of the system, and $p^z_{\text{swing}}, v^z_{\text{swing}}$ are the vertical position and velocity of the swing toe, respectively. This happens at the end of the SSP. Let $(q_k, \dot{q}_k)$ denote the pre-impact state at the step indexed by $k$. Then the state at next step $(q_{k+1}, \dot{q}_{k+1})$ at the switching surface can be represented by:
\begin{equation}
\label{XpartialMap}
   (q_{k+1}, \dot{q}_{k+1}) = \mathcal{P}(q_{k}, \dot{q}_{k}, \tau^{\ddot{L}},  u_k).
\end{equation}
In the literature of nonlinear dynamics \cite{sastry2013nonlinear}, Eq. \eqref{XpartialMap} is typically referred as the {Poincar\'e} return map. Here we denote it as the step-to-step (S2S) dynamics, i.e., the dynamics at the discrete step level. The intuitive description of Eq. \eqref{XpartialMap} is that the final state at the current step is affected by the state at the previous step and the actuation applied at the previous step. In terms of walking, it is important to make sure that the \textit{horizontal state} $\mathbf{x}^{\text{aSLIP}} = [p_x,  \dot{p}_x,p_y, \dot{p}_y]^T$ behaves appropriately. For an existing walking with the leg length trajectory remaining unchanged, perturbing the step size $u$ can potentially regulate the behavior of the horizontal states during walking. The same philosophy has also appeared in the capture point approach \cite{pratt2012capturability} and the Raibert controller \cite{raibert1986legged} where the step location is used for controlling robot balancing. Thus, the problem left here is how to control the horizontal states to exert desired behaviors by the changes of step size via the S2S dynamics. 

\subsection{Stepping Control via S2S Approximation}
Now we explain the stepping control via the approximation of the S2S dynamics in the integration with the periodic leg length actuation. By viewing the difference between the actual S2S dynamics and the approximation as the disturbance to the approximation dynamics, the system state $\mathbf{x}$ can be controlled so that the difference between $\mathbf{x}$ and the state of the approximation dynamics becomes bounded. 

The S2S dynamics of the horizontal states is given by,
\begin{equation}
\label{S2Strans}
    \mathbf{x}^{\text{aSLIP}}_{k+1} =  \mathcal{P}_{\mathbf{x}}(q_{k}, \dot{q}_{k}, \tau^{\ddot{L}},  u_k) ,
\end{equation}
where $\mathbf{x}^{\text{aSLIP}}_{k+1}$ is the horizontal state, and $\mathcal{P}_{\mathbf{x}} $ is the corresponding rows of the $\mathcal{P}$ in Eq. \eqref{XpartialMap}. The exact formulation of the S2S can not be found in closed-form. Here we will use the S2S dynamics of the H-LIP as the approximation (see Fig. \ref{fig:aSLIP} (b,c)), which is a linear system:
\begin{equation}
\label{linearReturnMap}
\mathbf{x}^{\text{H-LIP}}_{k+1} = A \mathbf{x}^{\text{H-LIP}}_k  +  B u^{\text{H-LIP}}_{k},
\end{equation}
where $\mathbf{x}^{\text{H-LIP}}$ and $u^{\text{H-LIP}}_{k}$ are the horizontal state and the step size of the H-LIP, respectively.
The exact derivation of its S2S dynamics will be illustrated in the next section. With the approximation, Eq. \eqref{S2Strans} can be rewritten as: 
\begin{align}
\label{disturbedReturnMap}
\mathbf{x}^{\text{aSLIP}}_{k+1} &= A \mathbf{x}^{\text{aSLIP}}_{{k}}  +  B u_{k} + w\\
 \label{modelDiff} 
 w &:= \mathcal{P}_{\mathbf{x}}(q_{k}, \dot{q}_{k}, \tau^{\ddot{L}},  u_k) - A \mathbf{x}^{\text{aSLIP}}_{{k}}  -  B u_{k},
\end{align}
where $w$ represents the difference of the S2S dynamics between the aSLIP and the H-LIP. If $\mathbf{x}^{\text{aSLIP}}_k =\mathbf{x}^{\text{H-LIP}}_{{k}}$ and $u_k =u^{\text{H-LIP}}_{{k}}$, then $\mathbf{x}^{\text{aSLIP}}_{k+1} = \mathbf{x}^{\text{H-LIP}}_{{k+1}} + w$. As the leg length of the aSLIP periodically retracts and extends, each step is assumed to happen in finite time and thus the S2S dynamics exists. The continuous dynamics of the aSLIP and the H-LIP themselves are bounded for walking with finite velocities. Thus $w$, the integration of the continuous dynamics error between the two over the step, is assumed to be bounded, i.e., $w \in W$. 

Suppose $u^{\text{H-LIP}}$ is designed to realize the desired walking behaviors on the H-LIP. Let $\mathbf{e}_k = \mathbf{x}^{\text{aSLIP}}_{k} - \mathbf{x}^{\textrm{H-LIP}}_{{k}}$ be the difference on the discrete horizontal state between the aSLIP and the H-LIP. 
 The \textbf{H-LIP based stepping controller}: 
\begin{equation}
\label{steppingController}
    u_k = u_k^{\text{H-LIP}} + K (\mathbf{x}^{\text{aSLIP}}_k -\mathbf{x}^{\text{H-LIP}}_{{k}}  )
\end{equation}
yields the closed-loop system of the aSLIP to be:
\begin{equation}
\label{aSLIPclosedloop}
    \mathbf{x}^{\text{aSLIP}}_{k+1} = A \mathbf{x}^{\text{aSLIP}}_{{k}}  +  B (u^{\text{H-LIP}}_k + K (\mathbf{x}^{\text{aSLIP}}_k -\mathbf{x}^{\text{H-LIP}}_{{k}}  )) + w.
\end{equation}
Subtracting Eq. \eqref{aSLIPclosedloop} by Eq. \eqref{linearReturnMap} on both sides of the equality sign, we have the \textit{error dynamics}: 
\begin{equation}
\label{errordynamics_cl}
    \mathbf{e}_{k+1} = (A + BK) \mathbf{e}_{k}  + w:=A_{cl} \mathbf{e}_{k}  + w.
\end{equation}
The error dynamics has a minimum disturbance invariant set $E$ if $A_{cl}$ is stable: 
\begin{equation}
\label{setE}
    A_{cl} E  \oplus  W \in E,
\end{equation} 
where $\oplus$ is the Minkowski sum. This means that if $\mathbf{e}_{k} \in E$, then $\mathbf{e}_{k+1} \in E$. In other words, if the state difference between the aSLIP and the H-LIP is initially bounded by $E$, then the difference will be always bounded by $E$ by applying the stepping controller in Eq. \eqref{steppingController}. Therefore, to achieve desired walking, we first design the nominal behavior $\mathbf{x}^{\text{H-LIP}}$, $u^{\text{H-LIP}}$ on the H-LIP, and then apply the stepping controller to find the desired step size on the aSLIP to generate an approximate behavior. We will further describe this in the next two sections, where we omit the superscript $^\text{H-LIP}$ when we describe the calculation of the nominal behavior on the H-LIP.

\section{The Hybrid Linear Inverted Pendulum Model}
\label{section:LIP}
In this section, we briefly describe the Hybrid Linear Inverted Pendulum (H-LIP) model, its periodic orbits, its step-to-step (S2S) dynamics, and the feedback stabilization. The purpose is to set up all the necessities for applying the H-LIP based stepping controller of Eq. \eqref{steppingController} in Section \ref{section:walking}. 
\subsection{Dynamics of walking} 
The H-LIP has a point mass on telescopic legs (Fig. \ref{fig:aSLIP} (c)). The point mass has a constant center of mass (COM) height. The  walking dynamics of the H-LIP have two alternating parts:
\begin{align}
 \ddot{p} &= \lambda^2 p,   \tag{SSP}  \\
 \ddot{p} &= 0,  \tag{DSP}
\end{align}
where $p$ is the mass forward position relative to the stance foot, $\lambda = \sqrt{\frac{g}{z_0}}$, and $z_0$ is the nominal height of the H-LIP. The dynamics in SSP is a passive LIP model. In DSP, the velocity of the mass is constant, which has also been assumed in \cite{kajita2002realtime, feng2016robust}. Additionally, we assume that the walking has constant domain durations, i.e., $T_\textrm{SSP}, T_\textrm{DSP}$ are constants. The transitions between the domains have no velocity jumps: 
\begin{equation}
 \Delta_{\text{SSP} \rightarrow \text{DSP}} : 
\left \{\begin{matrix}
v^{+} = v^{-}  \\
p^{+} = p^{-}
\end{matrix}, \right.  \ \
\Delta_{\text{DSP} \rightarrow \text{SSP}} :  \left \{\begin{matrix}
v^{+} = v^{-}  \\
p^{+} = p^{-}   - u
\end{matrix} \right . \nonumber
\end{equation}
where the superscripts $+/-$ indicate the beginning and the end of the domain, and $u$ is the step size. 

\subsection{Periodic Orbits:}
\begin{figure}[t]
      \centering
      \includegraphics[width = 1\columnwidth]{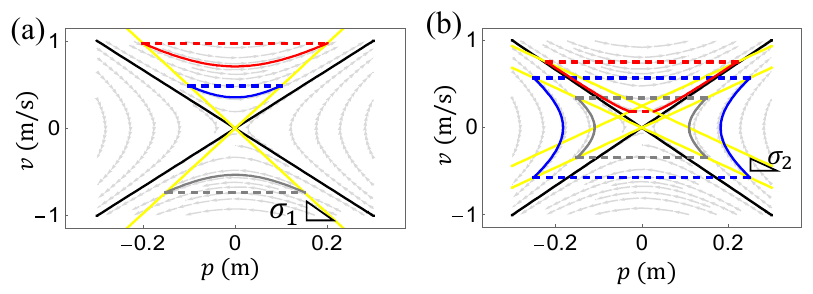}
      \caption{Examples of Period-1 (a) and Period-2 (b) orbits (indicated by red, blue and grey trajectories) in the phase portrait of SSP of the H-LIP, where the yellow lines are the characteristic lines.}
      \label{fig:orbits}
\end{figure}

Periodic walking of the H-LIP can be geometrically identified from its phase portrait in SSP in \cite{xiong2019IrosStepping}, where the orbits are classified into Period-1 (P1) orbits and Period-2 (P2) orbits (see Fig. \ref{fig:orbits}). Moreover, periodic orbits can be explicitly found in closed-form. Here we briefly present the procedure of orbit identification to realize the desired walking behaviors.

P1 orbits are the orbits that are realized via one single step. It is proven in \cite{xiong2019IrosStepping} that the boundary states of the SSP of all P1 orbits are on the characteristic lines $v = \pm\sigma_1 p$, where $\sigma_{1} = \lambda \textrm{coth} \left(\frac{T_\textrm{SSP}}{2} \lambda\right).
$
Given a desired velocity $v^d$, the realization with the P1 orbit is unique. The preimpact (final) state of SSP and the nominal step size of the P1 orbit are:
\begin{equation}
\label{P1nominal}
    [p^*, v^*]  = [1, \sigma_1]\frac{v^d T}{2 + T_\textrm{DSP} \sigma_1},\ \ u^* = v^d T,
\end{equation}
where $v^d$ is the desired net velocity, and $T = T_\textrm{DSP} +T_\textrm{SSP}$.

For P2 orbits, \cite{xiong2019IrosStepping} proved that all boundary states of the SSP are on the characteristic line $v = \pm\sigma_2 p + d_2$, where
\begin{equation}
\sigma_{2} = \lambda {\textrm{tanh} \left(\frac{\lambda}{2} T_\textrm{SSP} \right)}, \ \ 
d_2 = \frac{\lambda^2 \textrm{sech}^2 (\frac{\lambda}{2}T_{\textrm{SSP}}) v^{d} T}{\lambda^2 T_{\textrm{DSP}} + 2\sigma_{2}}.
\end{equation}
There are infinite number of equivalent P2 orbits to realize the same desired velocity. The difference among all the P2 orbits is on the step size. As each P2 orbit has two alternating step sizes, the sum of two consecutive ones is constant: 
\begin{equation}
\label{P2nominal}
    u^*_\text{L} + u^*_\text{R} = 2v^d T,
\end{equation}
where $u^*_\text{L/R}$ is the step size when the left/right foot is the original stance foot. Selecting either $u^*_\text{L}$ or $u^*_\text{R}$ determines the P2 orbit. The boundary states of SSP are thus determined, 
\begin{align}
\label{P2setP}
p^*_\text{L/R} = \frac{u^*_\text{L/R} - T_{\textrm{DSP}} d_2}{2 + T_{\textrm{DSP}} \sigma_2}, \ \ 
v^*_\text{L/R} = \sigma_2 p^*_\text{L/R} + d_2.
\end{align}
The P2 orbit is thus identified from its boundary states of the SSP and its corresponding step sizes. 

\subsection{S2S Dynamics and Stabilization}
The dynamics of the H-LIP is piece-wise linear. Since the durations are fixed, we can derive the discrete step-to-step (S2S) dynamics in closed-form where the step size $u$ becomes the control input. By selecting the final state of the SSP as the state of the discrete S2S dynamics, we have: 
\begin{equation}
\label{LIPS2S}
\mathbf{x}_{k+1} = A \mathbf{x}_{k} +
B u_k,
\end{equation}
where $\mathbf{x}_{k+1} = [p_{k+1}, v_{k+1}]^T$ is the current, $\mathbf{x}_{k} = [p_{k}, v_{k}]^T$ and $u_k$ are the state and the step size at previous step, and,
\begin{eqnarray}
\label{A}
A &= & \begin{bmatrix}
 \mathrm{cosh}(T_{\textrm{SSP}} \lambda) & T_{\textrm{DSP}} \mathrm{cosh}(T_{\textrm{SSP}} \lambda) + \frac{1}{\lambda}\mathrm{sinh}(T_{\textrm{SSP}}\lambda)\\ 
\lambda \mathrm{sinh}(T_{\textrm{SSP}} \lambda) &  \mathrm{cosh}(T_{\textrm{SSP}} \lambda) + T_{\textrm{DSP}} \mathrm{sinh}(T_{\textrm{SSP}} \lambda)
\end{bmatrix},  \nonumber \\
\label{B}
B  &= & \begin{bmatrix}
 - \mathrm{cosh}(T_{\textrm{SSP}} \lambda)\\ 
- \lambda \mathrm{sinh}( T_{\textrm{SSP}} \lambda)
\end{bmatrix}.\nonumber
\end{eqnarray}
Therefore, Eq. \eqref{LIPS2S} is the S2S dynamics of the H-LIP. 

In the discrete S2S dynamics formulation, individual P1 orbits are represented by the set point $\mathbf{x}^* = [p^*, v^*]^T$; P2 orbits are represented by the alternating set point $\mathbf{x}^*_\text{L/R} = [p^*_\text{L/R}, v^*_\text{L/R}]^T$. Orbit stabilization can be derived based on the linear S2S dynamics in Eq. \eqref{LIPS2S}. Without further illustrations, it can be easily proven that both P1 and P2 orbits can be stabilized respectively using the controllers:
\begin{align}
\label{P1stb}
    u_1 = K( \mathbf{x} - \mathbf{x}^*) + u^*, \ \ 
     u_2 = K( \mathbf{x} - \mathbf{x}^*_\text{L/R}) + u^*_\text{L/R},
\end{align}
where $K$ is the feedback gain to make the closed-loop matrix $A_{cl} = A + BK$ stable. Note that the controllers are also in the form of Eq. \eqref{steppingController}. One can view the orbit stabilization of the H-LIP as stabilizing the H-LIP that is not on the orbit yet to another H-LIP that is evolving on the orbit. Since the dynamics difference is zero, the error will be driven to zero.  

When $K$ is the deadbeat gain, Eq. \eqref{P1stb} can yield identical stepping controllers in Theorem 2.1 and 2.2 in \cite{xiong2019IrosStepping}. Since the system has two states and one input, it requires two steps to drive the error to zero. Thus the deadbeat gain $K_{\text{deadbeat}}$ is calculated by solving:
$
    (A+BK_{\text{deadbeat}})^2 = 0.
$
Additionally, the linear quadratic regulator (LQR) provides a family of gains which minimizes a custom cost function:
\begin{equation}
\label{LQRcost}
J_{\text{LQR}} =\textstyle \sum_{k = 1}^{\infty} ( \mathbf{x}_k^T Q \mathbf{x}_k + u_k^T R u_k + 2\mathbf{x}_k^T N u_k).
\end{equation}
The resultant optimal state feedback gain is:
\begin{equation}
\label{LQRgain}
K_{\text{LQR}} = -(R + B^T P B)^{-1}(B^T P A + N^T),
\end{equation}
where $P$ is solved from the discrete-time algebraic Riccati equation. Both $K_{\text{deadbeat}}$ and $K_{\text{LQR}}$ can be used in Eq. \eqref{P1stb} for stabilization and in Eq. \eqref{steppingController} for controlling aSLIP walking. 

\section{Dynamic Walking Generation of aSLIP via H-LIP Stepping}
\label{section:walking}

Now we apply the H-LIP based stepping controller (Eq. \eqref{steppingController}) for generating versatile walking on the aSLIP. The H-LIP was described in the planar case, and 3D walking of the aSLIP is realized via an orthogonal composition of two H-LIPs. We describe the generation of periodic walking first and then the versatile walking using constrained optimization on the H-LIP.   

\subsection{3D Periodic Walking Generation}

To realize 3D periodic walking on aSLIP, we first select and compose P1 and P2 orbits of the H-LIP into 3D. Two types of orbits in two planes ($x-z$ and $y-z$) provide three possible combinations, i.e., two P1 orbits in 3D (P1-P1), two P2 orbits (P2-P2), and one P1 and one P2 orbit (P1-P2). One important concern towards the embedding on the humanoid is that the step sizes have to be larger than the foot sizes to avoid kinematic violations. Thus P1-P1 is not possible to realize; step sizes are 0 for zero speed walking, which creates violations. P2-P2 and P1-P2 can be realized. 

We illustrate the process by generating a forward walking with a P1-P2 composition. A P1 orbit is selected in the $x-z$ plane with $v^d = 0.3$m/s. The desired set point $\mathbf{x}^*$ of the S2S dynamics in $x-z$ plane and the nominal step size $u^*$ are calculated in Eq. \eqref{P1nominal}. A P2 orbit is selected in the $y-z$ plane with $v^d = 0$m/s. Then we select the step size $u^*_\text{L}$ to be larger than the foot width of the humanoid for avoiding foot collision, and $u^*_\text{R}$ is calculated using Eq. \eqref{P2nominal}. The set point $\mathbf{x}^*$ in the $y-z$ plane is calculated in Eq. \eqref{P2setP}. The desired walking behaviors of the H-LIPs are thus identified, which will be used in the stepping controller in Eq. \eqref{steppingController}. 

\begin{figure}[t]
      \centering
      \includegraphics[width = 1\columnwidth]{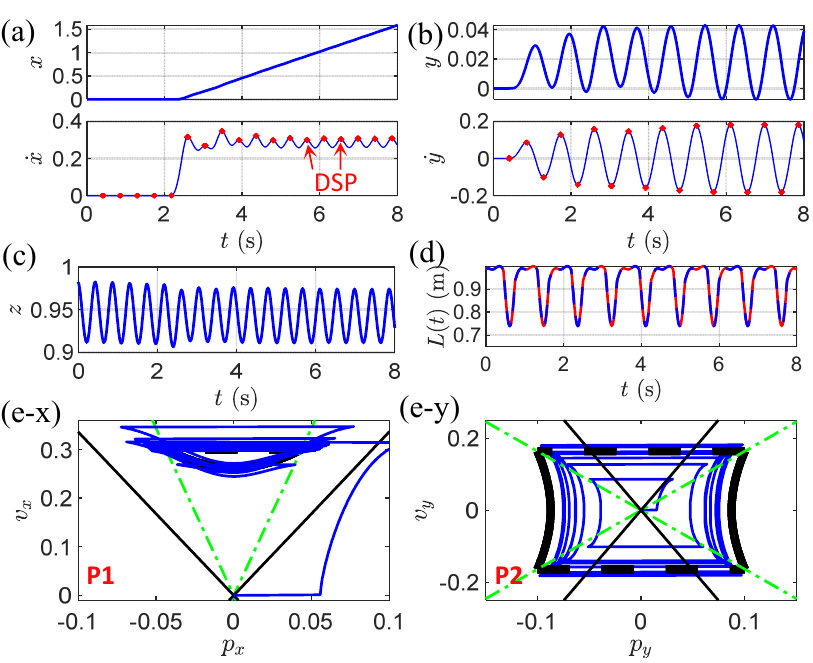}
      \caption{An example of the generated 3D periodic walking: (a-b) horizontal trajectories of the mass vs time (the red indicates the velocities in DSP), (c) the vertical trajectory of the mas, (d) the leg length tracking (the red and blue are the desired and actual $L$, respectively), and (e) the phase portraits of the mass (blue) in the $x-z$ plane (e-x) and $y-z$ plane (e-y) comparing to the desired orbits of the H-LIP (black).}
      \label{fig:3Dperiodic}
\end{figure}

Let the aSLIP start from stepping-in-place, 
The average height of the aSLIP is selected as $z_0$ for the H-LIP. The spring parameters are selected as $K_s = 24000$N/m, and $D_s = 700$Ns/m. The vertical oscillation of the mass is about 5cm, which can be tuned from the trajectory optimization. The optimized leg length is selected as the desired periodical leg length $L^{\text{des}}(t)$. Eq. \eqref{leglengthActuation} is applied to control the leg length. Executing the step size from the stepping control in Eq. \eqref{steppingController} perturbs the stepping-in-place. Fig. \ref{fig:3Dperiodic} shows the simulation result. The aSLIP is commanded to walk forward at $t = 2$s. As indicated in the phase plot in Fig. \ref{fig:3Dperiodic} (c-xy), the horizontal states of the aSLIP converge closely to the desired P1 and P2 orbits of the H-LIP in each plane. The 3D periodic walking is thus generated on the aSLIP.

\subsection{Versatile Walking via Constrained Optimal Control} 
The desired periodic walking behavior and corresponding nominal step size of the H-LIP are in closed-form. For other versatile walking behaviors, constrained optimization problems can be formulated on the linear S2S dynamics of the H-LIP to find the desired H-LIP state sequence and the nominal step size sequence. Then the stepping controller can be applied on the aSLIP to realize similar behaviors. 

For non-periodic walking, it is desirable to define the \textit{extended horizontal state} $\tilde{\mathbf{x}}$ in the S2S dynamics to include the global position $x$ of the point mass. 
The planar S2S dynamics of the H-LIP with $\tilde{\mathbf{x}}$ can be described as: 
\begin{equation}
\label{LIPglobalLTI}
\underset{\tilde{ \mathbf{x}}_{k+1}}{\underbrace{\begin{bmatrix}
x_{k+1}\\
p_{k+1}\\
v_{k+1}
\end{bmatrix}}} = \underset{\tilde{A}}{\underbrace{\begin{bmatrix}
1 & A_{1,1} - 1 & A_{1,2} \\
0 & A_{1,1} & A_{1,2}\\
0 & A_{2,1} & A_{2,2}
\end{bmatrix}}}
\underset{\tilde{\mathbf{x}}_{k}}{\underbrace{\begin{bmatrix}
x_{k}\\
p_{k}\\
v_{k}
\end{bmatrix}}} +
\underset{\tilde{B}}{\underbrace{\begin{bmatrix}
B_{1} + 1\\
B_{1}\\
B_{2}
\end{bmatrix}}} u_k, \nonumber
\end{equation}
where the subscripts indicate the entries of $A, B$ in Eq. \eqref{LIPS2S}.

The desired behavior can be directly described on the states via the cost function or constraints in the formulation of convex optimizations to optimize a sequence of step sizes. Since the S2S dynamics is linear, the optimization is formulated as a quadratic program (QP), which is fast to be solved. The optimization is typically defined as: 
\begin{align}
\label{MPC-LIP}
u_{1, \dots, N}, \tilde{\mathbf{x}}_{1, \dots, N} = & \underset{ \{u_{1, \dots, N}, \tilde{\mathbf{x}}_{1, \dots, N} \} \in \mathbb{R}^{4 \times N}    }
 {\text{argmin}}  J_\text{QP} \\
\text{s.t.} \quad & \tilde{\mathbf{x}}_{k+1} = \tilde{A} \tilde{\mathbf{x}}_k + \tilde{B} u_k , k = \{1, \dots, N-1\}\nonumber\\
\quad &  \left |  u_k \right | < u_{\text{max}} , k = \{1, \dots, N\}  \nonumber \\
  \quad &  \tilde{\mathbf{x}}_1 = \tilde{\mathbf{x}}_0 \  \nonumber
\end{align}
where $J_{\text{QP}}$ is the task-dependent cost function, $u_{1, \dots, N}$ are the optimized step sequence, $\tilde{\mathbf{x}}_{1, \dots, N}$ are the optimized states of the H-LIP, and $\tilde{\mathbf{x}}_0$ is the initial state of the H-LIP. Here the planar case is presented. For the 3D case, it is a composition of two QPs of planar problems into one QP. 

The solution of the QP provides the behavior of the H-LIP in the stepping controller in Eq. \eqref{steppingController} for achieving a similar behavior on the aSLIP. The difference between the horizontal state of the aSLIP and that of the H-LIP is bounded by the disturbance invariant set in Eq. \eqref{setE}. Here we demonstrate the approach in the following scenarios of walking.

\subsubsection{Fixed location tracking}
We suppose the task is to walk to a fixed location, which oftentimes is a practical task on the robot. Using the constrained optimization approach, the task can be encoded as an equality constraint: $\tilde{\mathbf{x}}_N = \tilde{\mathbf{x}}^d$ or as a part of the cost: 
 $ J_\text{QP} = \textstyle \sum_{k = 1}^{N} (\tilde{\mathbf{x}}_k^T -   \tilde{\mathbf{x}}^d)^T Q(\tilde{\mathbf{x}}_k^T -   \tilde{\mathbf{x}}^d) +u_k^T R u_k$,
where $Q, R$ are the parameters of the cost. Here we demonstrate the approach by controlling the aSLIP to a forward location in the $x-z$ plane where $\tilde{\mathbf{x}}^d = [1, 0, 0]^T$. The optimized result on the H-LIP and the generated walking on the aSLIP are shown in Fig. \ref{fig:pathSLIP}. The aSLIP walks to the desired location and stays there. In the $y-z$ plane, a P2 orbit is applied to keep the feet apart on the ground.

\subsubsection{Trajectory Tracking}
\begin{figure}[t]
      \centering
      \includegraphics[width = 1\columnwidth]{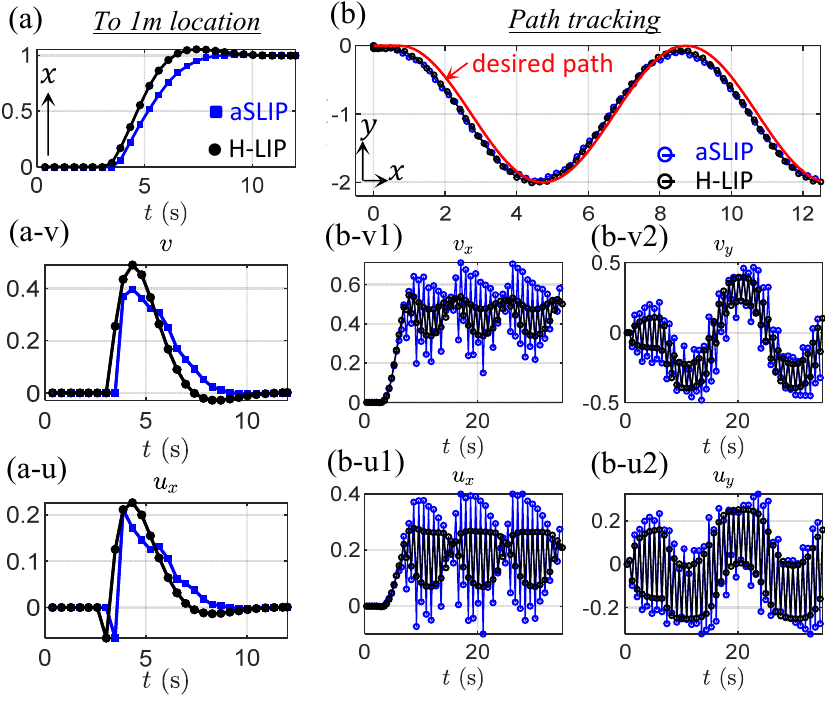}
      \caption{(a) Results of the fixed location tracking in terms of the forward positions $x$, forward velocities $v$ and step sizes $u_x$. (b) Results of the path tracking with the velocities and step sizes in each plane. The black and blue indicate the results of the H-LIP and the aSLIP, respectively.}
      \label{fig:pathSLIP}
\end{figure}

Now we consider the task is to track a trajectory on the ground. We assume that the trajectory is provided by a high-level planner using sensors on the humanoid. Let $\tilde{\mathbf{x}}^d(t)$ be the desired trajectory w.r.t. time. The tracking is directly encoded via the cost: 
\begin{equation}
    J_\text{QP} = \textstyle \sum_{k = 1}^{N} (\tilde{\mathbf{x}}_k^T -   \tilde{\mathbf{x}}^d(kT))^T Q(\tilde{\mathbf{x}}_k^T -   \tilde{\mathbf{x}}^d(kT)) +u_k^T R u_k. \nonumber
\end{equation}
The QP can be solved recursively in Model Predictive Control (MPC) fashion since in practice the desired trajectory may be available only in a short future horizon. Since the trajectory tracking is in 3D, the QP is formulated including states in both $x-z$ and $y-z$ planes. An additional constraint is that the step size should be larger than the foot size of the humanoid. Here we apply the method to track a sinusoidal path, shown in Fig. \ref{fig:pathSLIP}, where the aSLIP behaves closely to the H-LIP and tracks the desired trajectory closely. 

\textit{Remark:} The constrained optimization on the H-LIP model shows similarities to the LIP-ZMP methods \cite{diedam2008online, feng2016robust} where the LIP model admits quadratic programs for planning both ZMP and step size. The step planning here can be viewed as the special case that the foot actuation is zero. However, one should note that the planned motion of the H-LIP is applied not directly on the humanoid, but in the H-LIP based stepping controller for the aSLIP.

\begin{figure}[t]
      \centering
      \includegraphics[width = 0.9\columnwidth]{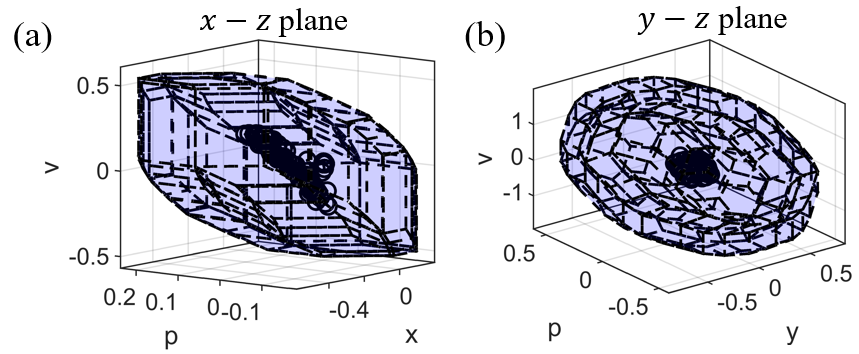}
      \caption{The disturbance invariant set $E$ in blue transparent polytopes and the state difference $\mathbf{e}$ in black dots each plane with $K = [0.21,0.96,0.52]$ in the $x-z$ plane and $K = [0.31,0.67,0.43]$ in the $y-z$ plane.}
      \label{fig:analysis}
\end{figure}

\subsection{Analysis}
The simulation results on the aSLIP indicate the success of applying the stepping controller in Eq. \eqref{steppingController}. One thing to note is that the initial H-LIP state should be close to that of the horizontal state of the aSLIP, such that $\mathbf{e}_0  = \tilde{\mathbf{x}}^{\text{aSLIP}}_0 - \tilde{\mathbf{x}}^{\text{H-LIP}}_0 \in E $. In the examples, we select the H-LIP state to be identical to that of the aSLIP in the beginning, which indicates $\mathbf{e}_0 = \mathbf{0} \in E $. Then $\mathbf{e}_k \in E$ for all $k\in \mathbb{N}$ under the closed-loop system in Eq. \eqref{errordynamics_cl}. To verify this, we first numerically calculate $w$ in Eq. \eqref{modelDiff} in the simulated walking, and then approximate $W$ via a polytope. When $K_\text{deadbeat}$ is used, $A_{cl}$ is nilpotent ($\exists n$, s.t. $(A_{cl})^n = \mathbf{0}$). Thus $E = E_n := \oplus_{i=0}^{n-1} A^i_{cl} W$ can be calculated exactly \cite{rakovic2005invariant}. When $K_\text{LQR}$ is used, $E$ can be inner-approximated with $E_{n \in \mathbb{N}}$ (i.e. $E_n \subset E$) or outer-approximated using the techniques in \cite{rakovic2005invariant} and the reference therein. We apply set iterations using MPT \cite{MPT3} to calculate $E_{n=6}$ as the inner approximation. Fig. \ref{fig:analysis} shows the approximation of $E$ and the error $\mathbf{e}$ for the example of trajectory tracking. The different LQR gains in each plane come from a different $Q$, $R$ in Eq. \eqref{LQRgain}. $\mathbf{e} \in E$, which verifies the application of the stepping controller that keeps the error small. 

\section{Versatile Walking Embedding}
\label{section:robot}

\begin{figure}[t]
      \centering
      \includegraphics[width = 0.85\columnwidth]{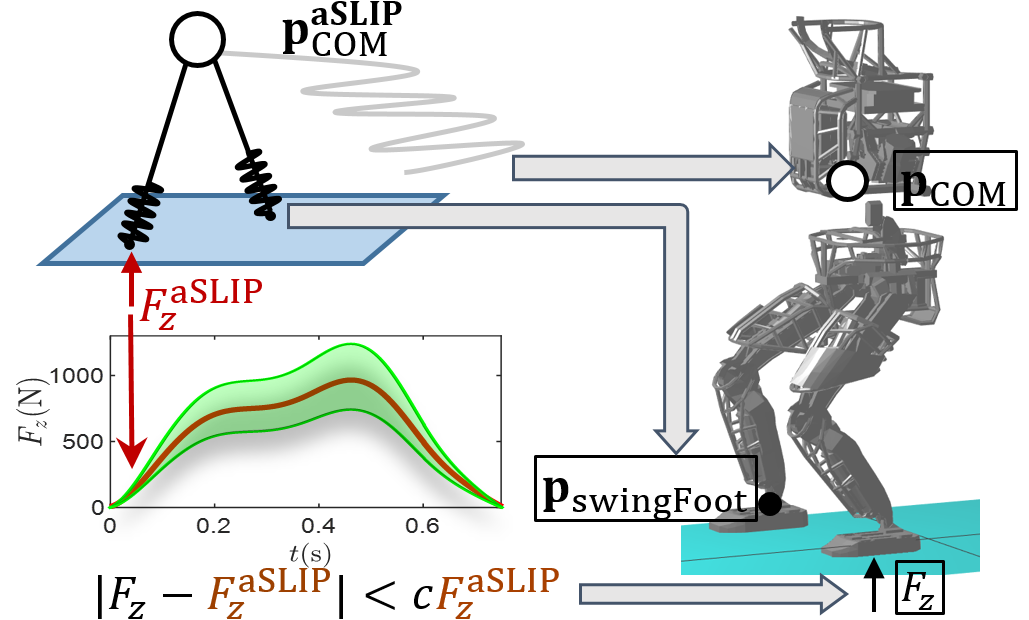}
      \caption{The dynamics embedding scheme on Atlas.}
      \label{fig:embedding}
\end{figure}

In this section, we describe the embedding of the aSLIP walking on the humanoid via the COM, step locations, and contact forces (see Fig. \ref{fig:embedding}). The robot model of Atlas \cite{feng2016robust} \cite{atlasurdf} (without arms) is used. We first describe the dynamics model of the walking, the output definition, and finally the output stabilization via the task space controller. 

\subsection{Robot Model of Walking}
Let $q_r$ describe the robot configuration with floating base coordinates. The continuous dynamics of walking is described by the Euler-Lagrange equation:
\begin{eqnarray}
&& M\ddot{q}_r + h = B \tau + J_{v}^T F_{v},  \label{eom} \\
&& J_{v} \ddot{q}_r + \dot{J}_{v} \dot{q}_r = 0, \label{hol}
\end{eqnarray}
where $M$ is the mass matrix, $h$ is the Coriolis, centrifugal and gravitational term, $B$ is the actuation matrix, $\tau \in \mathbb{R}^{12}$ is the motor torque vector, and $F_v$ and $J_v$ are the contact force vector and its corresponding Jacobian, respectively.
Depending on the number of feet in contact with the ground, the walking is composed of the SSP and DSP. 

\textbf{Contact Model:} We assume that the ground is rigid and that the impact is plastic \cite{grizzle2014models} when the foot strikes the ground. For simplification, the foot is assumed to land on the ground flatly. The velocity of the robot undergoes a discrete jump at the impact event, which is written as:
$\dot{q}_{r_{\text{DSP}}}^{+} = \Delta(q_r)\dot{q}_{r_{\text{SSP}}}^{-}.
$

\subsection{Output Construction}
The walking behavior is described via trajectories of the outputs, i.e, the important features. For DSP, two feet are constrained on the ground. The outputs are defined as:
\begin{equation}
\label{eq:outputDSP}
 \mathcal{Y}_{\textrm{DSP}}(q_r,t) = \begin{bmatrix}
  \mathbf{p}_{\textrm{COM}}(q_r) \\
    \pmb{\phi}_{\textrm{pelvis}}(q_r)
     \end{bmatrix} -
\begin{bmatrix}
 \mathbf{p}^{\text{aSLIP}}_{\textrm{COM}}(t) \\
 \mathbf{0}  \end{bmatrix},
\end{equation}
which includes the position of the COM and the orientation of the pelvis. The desired pelvis orientation is fixed, and the desired COM position is the aSLIP mass trajectory.

For SSP, the swing foot position $\mathbf{p}_{\textrm{swingFoot}}(q_r)$ and orientation $\pmb{\phi}_{\textrm{swingFoot}}(q_r)$ are controlled as well. $\pmb{\phi}_{\textrm{swingFoot}}(q_r)$ is fixed, and $\mathbf{p}_{\textrm{swingFoot}}(q_r)$ is constructed towards the desired step location of the aSLIP walking. Thus the outputs of SSP are:
\begin{equation}
\label{eq:outputSSP}
 \mathcal{Y}_{\textrm{SSP}}(q_r,t) = \begin{bmatrix}
\mathbf{p}_{\textrm{COM}}(q_r) \\
  \pmb{\phi}_{\textrm{pelvis}}(q_r)  \\
    \mathbf{p}_{\textrm{swingFoot}}(q_r) \\
     \pmb{\phi}_{\textrm{swingFoot}}(q_r)   \end{bmatrix} -
\begin{bmatrix}
 \mathbf{p}^{\text{aSLIP}}_{\textrm{COM}}(t) \\
 \mathbf{0}\\
   \mathbf{p}^{\text{desired}}_{\textrm{swingFoot}}(t) \\
 \mathbf{0}   \end{bmatrix}.
\end{equation}

\subsection{Output Stabilization with Active Force Control}
To drive the outputs to zero, we apply the
active contact force control in \cite{xiong2019exo} with the task space  controller (TSC) \cite{sadeghian2013task,wensing2013generation}. By encoding the desired output acceleration (second-order derivatives) in the cost, the optimized torques and contact forces are smooth. First, differentiating the output $\mathcal{Y}$ twice yields, 
\begin{equation}
    \ddot{\mathcal{Y}} = f_{\mathcal{Y}}(q_r, \dot{q}_r) + g_{\mathcal{Y}}(q_r, \dot{q}_r) \tau.
\end{equation}
Selecting desired $\ddot{\mathcal{Y}}^{des} = K_p \mathcal{Y} + K_d \dot{\mathcal{Y}}$ yields a desired exponential stable linear system.
An optimization problem can be formulated to solve for optimal $\tau$ by minimizing:
\begin{equation}
    J_{\textrm{TSC}}(\tau) = || \ddot{\mathcal{Y}} - \ddot{\mathcal{Y}}^{des} ||,
\end{equation}
which is a quadratic cost on $\tau$. Additionally, the input has to satisfy the max torque constraint: $|\tau| < \tau_{\text{max}}$ and the constraints on the ground contact forces (GRF): $R_F F_v < 0$, where $R_F$ is a constant matrix that encodes the friction cone and the non-negative normal forces. Solving $F_v$ from Eq. \eqref{eom} and \eqref{hol} yields the affine relation with respective to $\tau$, 
\begin{align}
 F_v &= A_v \tau + b_v, \label{utoF}
\end{align}
where the exact formulations of $A_v$ and $b_v$ are omitted. Thus the inequality on $F_v$ is equivalently an inequality on $\tau$.

\textbf{Contact Force Embedding:} To create smooth transitions between the two domains in the walking \cite{xiong2019exo}, we make sure that the Atlas walking exhibits similar ground normal reaction force profile to that of the aSLIP, i.e., $F_z \approx F_z^{\text{aSLIP}}$. This can be realized by setting up an inequality constraint, 
\begin{equation}
    (1-c) F_z^{\text{aSLIP}} < F_z < (1+c) F_z^{\text{aSLIP}}
\end{equation}
where $c\in (0,1)$ is an relaxation parameter. This inequality is translated on $\tau$ via Eq. \eqref{utoF}. 

\textbf{Quadratic Program:} The final controller to find optimal input $\tau$ is solving the quadratic program, defined as: 
\begin{align}
\tau^{*} &=  \underset{u \in \mathbb{R}^{12}} {\text{argmin}}  \quad J_{\textrm{TSC}} (\tau), \label{eq:TSC} \\
&\text{s.t.}  \quad   A_{\text{GRF}} \tau \leq b_{\text{GRF}},    \tag{GRF} \nonumber  \\
&  \quad  \quad \ \  |\tau| < \tau_{\text{max}}, \tag{Torque Limit}\nonumber
 \end{align}
 where the GRF constraint contains the above-mentioned constraints on the GRF. 
\subsection{Results}
\begin{figure}[t]
      \centering
      \includegraphics[width =1 \columnwidth]{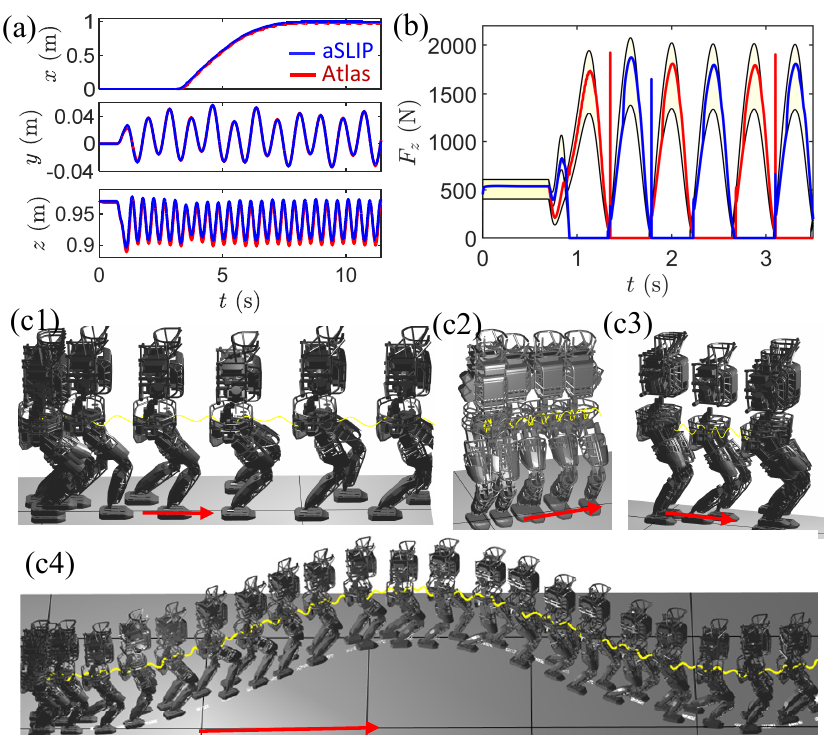}
      \caption{Simulation results: (a) the mass trajectory for the fixed location tracking, (b) the ground normal force during walking inside the relaxation, and the snapshots of the generate walking: (c1, c2) periodic walking in forward and lateral direction, and (c3, c4) tracking a fixed location and a global trajectory. }
      \label{fig:results}
\end{figure}

The embedding is realized in simulation \cite{Supplementary}. Fig. \ref{fig:results} shows the results. The force relaxation is selected as $c = 0.2$. We demonstrate the periodic walking, fixed location tracking, and trajectory tracking. Walking motions with different COM height variations are also generated. The walking of the aSLIP is embedded on Atlas via the output construction (Eq. \eqref{eq:outputDSP} \eqref{eq:outputSSP}) and the output stabilization using Eq. \eqref{eq:TSC}. For global trajectory tracking, we also enable turning on the robot by letting the pelvis and the swing foot to turn towards the walking direction. The simulated walking on Atlas verifies the proposed approach on walking generation via embedding the aSLIP walking with the H-LIP based stepping.

\section{Discussion and Future Work}
\label{section:discussion}
The proposed approach generates dynamic and versatile humanoid walking in a computation efficient way. This also motivates several research directions to discuss. 

The method assumes flat terrain for walking. Generating and stabilizing walking on uneven terrain can be challenging. Using the reduced order models (ROM), e.g., the aSLIP model, can presumably simplify the planning process. However, dramatically varying terrain height may lead the S2S approximation via H-LIP to fail. Step planning and other actuation planning may require more sophisticated care.  

Our method focuses on generating the desired COM dynamics. Cooperating that with the rotational dynamics may utilize all available degrees of freedom for controlling walking. For instance, a flywheel based SLIP model may also plan the rotational dynamics appropriately, and consequently, arms on the humanoid can be better used for walking. 

For the stepping controller, adaptive controllers can be explored to better stabilize the walking to the desired behaviors. Additionally, better approximations of the S2S dynamics can reduce the set $W$. Optimization on the gain $K$ in Eq. \eqref{steppingController} can further reduce the size of the disturbance invariant set $E$. The influence of the parameters of the aSLIP and the original stepping-in-place behavior on $W$ and then $E$ is also important to study. These will be our future work. 

\section{Conclusion}
\label{section:conclusion}
To conclude, we propose a novel and efficient approach to generate dynamic and versatile walking on humanoid robots. We generate target walking behaviors on the 3D actuated Spring Loaded Inverted Pendulum (3D-aSLIP) via the Hybrid Linear Inverted Pendulum (H-LIP) based stepping controller by perturbing an existing periodic walking of the 3D-aSLIP. The stepping controller is solved either in closed-form or via fast solvable quadratic programs. The generated 3D-aSLIP walking is then realized on the humanoid via an optimization-based controller with ground reaction force embedding. By innovating and exploiting the benefits of reduced order models, the approach renders dynamic and versatile walking generation on high-dimensional humanoids to be highly efficient. 


\bibliographystyle{IEEEtran}
\bibliography{Walking}

\end{document}